\documentclass[twocolumn, switch]{article} 

\usepackage{preprint}
\usepackage{algorithm}
\usepackage{algorithmic}
\usepackage{amsmath, amsthm, amssymb, amsfonts}

\usepackage[numbers,square]{natbib}
\usepackage[utf8]{inputenc}	
\usepackage[T1]{fontenc}	
\usepackage{xcolor}		
\usepackage[colorlinks = true,
            linkcolor = purple,
            urlcolor  = blue,
            citecolor = cyan,
            anchorcolor = black]{hyperref}	
\usepackage{graphicx}		
\usepackage{booktabs} 		
\usepackage{nicefrac}		
\usepackage{microtype}		
\usepackage{float}			

\usepackage{titlesec}
\titlespacing\section{0pt}{12pt plus 3pt minus 3pt}{1pt plus 1pt minus 1pt}
\titlespacing\subsection{0pt}{10pt plus 3pt minus 3pt}{1pt plus 1pt minus 1pt}
\titlespacing\subsubsection{0pt}{8pt plus 3pt minus 3pt}{1pt plus 1pt minus 1pt}

\title{REOPD: Reliability-Adaptive Reward Extrapolation \\ for On-Policy Distillation}

\usepackage{titling}
\author{%
\textbf{Yang Sun}\textsuperscript{1,*}\quad
\textbf{Lichao Ma}\textsuperscript{2,*}\quad
\textbf{Houyuan Qin}\textsuperscript{3,*}\\[0.25em]
\textbf{Yuxin Liu}\textsuperscript{4}\quad
\textbf{Hanyang Lu}\textsuperscript{5}\quad
\textbf{Yao Zhu}\textsuperscript{6}\\[0.25em]
\textbf{Pinlong Cai}\textsuperscript{1}\quad
\textbf{Guohang Yan}\textsuperscript{1}\\[0.6em]
\small
\begin{tabular}[t]{c}
\textsuperscript{1}Shanghai Artificial Intelligence Laboratory\\
\textsuperscript{2}Peking University\\
\textsuperscript{3}Southwest Jiaotong University\\
\textsuperscript{4}University of Science and Technology of China\\
\textsuperscript{5}University of Electronic Science and Technology of China\\
\textsuperscript{6}Renmin University of China\\[0.3em]
\textsuperscript{*}Equal contribution.\\
Email (Yang Sun): \texttt{25213080092@m.fudan.edu.cn}
\end{tabular}
}
\date{}

\begin{document}

\twocolumn[ 
  \begin{@twocolumnfalse} 

\maketitle
\thispagestyle{empty}

\begin{abstract}
On-policy distillation (OPD) trains a student on its own generated trajectories
under dense token-level supervision from a teacher, providing an effective
post-training paradigm for large language models. Reward-extrapolation methods
such as ExOPD further amplify the teacher--reference log-likelihood ratio to
move beyond direct imitation. However, ExOPD uses a single global scalar
$\lambda$ to apply the same extrapolation strength indiscriminately to every
token. This can drive the student to aggressively fit extreme peaks in the
teacher--reference log-ratio that defines the implicit reward, resulting in
reward hacking and unstable training. Moreover, the optimal $\lambda$ varies
across domains, requiring costly domain-specific sweeps that may still fail to
identify an appropriate extrapolation strength. We propose REOPD, a
Reliability-Adaptive Reward Extrapolation framework for On-Policy Distillation.
REOPD combines a token-level compatibility weight with a batch-level adaptive
budget. The former modulates token-wise residuals according to the
student--teacher discrepancy, while the latter dynamically adjusts the overall
extrapolation strength according to the reliability and scale of residual
signals in each batch. Together, they yield a token-wise effective coefficient
$\lambda_{b,t}=1+\gamma_b q_t$, which preserves the original teacher-alignment
term while selectively extrapolating along reliable teacher--reference
directions. REOPD requires no additional verifier, reward model, value model,
or rollout beyond the standard OPD pipeline. Evaluations show that REOPD
outperforms G-OPD on single-teacher mathematics and on both domains in the
multi-teacher setting, while matching G-OPD on single-teacher code. These
results demonstrate the effectiveness of fine-grained reliability adaptation
for reward extrapolation in on-policy distillation across task domains and
teacher configurations.
\end{abstract}
\vspace{0.35cm}

  \end{@twocolumnfalse} 
] 

\section{Introduction}
\label{sec:introduction}

On-policy distillation (OPD) trains a student on its own trajectories and
queries the teacher on the same prefixes \cite{agarwal2024onpolicy}. Unlike
distillation on fixed teacher-generated data, OPD supervises states that the
current student actually visits and provides a dense learning signal at every
sampled token. This makes it an effective post-training approach for
transferring reasoning capabilities from specialized teachers. Yet dense
supervision is heterogeneous: its usefulness depends not only on teacher
quality, but also on the local compatibility between student and teacher
\cite{li2026rethinking,xu2026tip}.

G-OPD interprets OPD as dense KL-regularized reinforcement learning, in which
the teacher--reference log-ratio defines an implicit token-level reward
\cite{yang2026learning}. Standard OPD corresponds to a reward coefficient
$\lambda=1$, whereas ExOPD sets $\lambda>1$ to extrapolate beyond direct
teacher matching. ExOPD, however, applies one global coefficient to every
token throughout training. Uniform scaling can overemphasize extreme
teacher--reference log-ratios together with useful residuals, allowing a small
number of peaks to dominate policy updates and increasing the risk of reward
hacking or unstable optimization. Moreover, the preferred coefficient differs
across domains, so each new setting requires multiple full training and
evaluation runs to select $\lambda$.

\begin{figure*}[t]
\centering
\includegraphics[width=\textwidth]{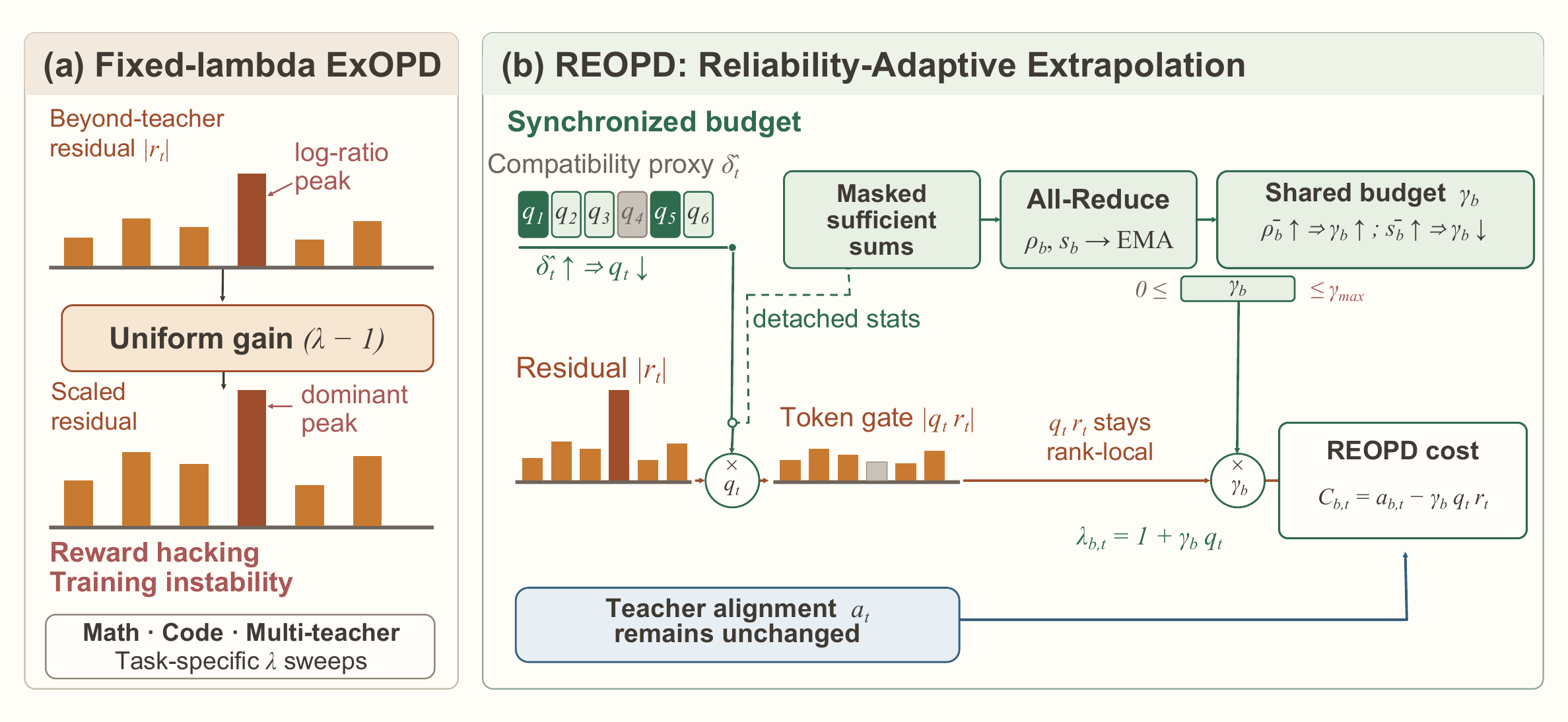}
\caption{Fixed-coefficient ExOPD applies the same residual gain
$\lambda-1$ to every token and may amplify teacher--reference log-ratio
peaks. REOPD preserves teacher alignment, gates the beyond-teacher residual
with compatibility $q_{b,i,t}$, and derives a bounded micro-batch budget
$\gamma_b$ from synchronized statistics.}
\label{fig:overview}
\end{figure*}

Our key observation is that \emph{teacher alignment} and \emph{extrapolation
beyond the teacher} should be controlled separately. A token with a large
student--teacher discrepancy may still provide a useful alignment signal, yet
be unsuitable for amplified extrapolation. We therefore propose REOPD, which
preserves the standard OPD alignment term and adapts only the additional
teacher--reference residual. As shown in Figure~\ref{fig:overview}, REOPD
combines a token-level compatibility weight with a bounded micro-batch budget.
Their product defines a token-wise effective coefficient, allowing reliable
residuals to be emphasized without uniformly increasing every update. The
method reuses the student, teacher, and reference log-probabilities already
available in G-OPD and requires no verifier, reward model, value model, or
additional rollout.

We evaluate REOPD on mathematical reasoning, code generation, and mixed-domain
multi-teacher distillation. Across repeated experiments, REOPD outperforms
G-OPD on single-teacher mathematics and on both mathematics and code in the
multi-teacher setting, while achieving comparable performance to G-OPD on
single-teacher code. These results indicate that online residual control can
reduce reliance on setting-specific coefficient sweeps while maintaining
competitive performance across task domains and teacher configurations.

Our contributions are summarized as follows:

\begin{itemize}
    \item We formulate the indiscriminate amplification and domain-specific
    tuning of fixed-coefficient reward extrapolation as a residual-control
    problem.

    \item We propose REOPD, which combines token-level compatibility with a
    bounded micro-batch budget while preserving the standard OPD alignment
    term and requiring no external outcome supervision.

    \item We evaluate REOPD against OPD and a fixed-coefficient sweep on
    mathematics, code, and multi-teacher distillation, showing competitive
    performance without selecting a task-specific global coefficient.
\end{itemize}

\section{Related Work}
\label{sec:related_work}

\paragraph{Knowledge Distillation and On-Policy Distillation.}
Knowledge distillation transfers teacher knowledge through output
distributions, representations, or generated sequences
\cite{hinton2015distilling,kim2016sequence}. Sequence-level distillation for
autoregressive models is usually off-policy, creating a mismatch between
teacher trajectories used for training and student prefixes encountered at
inference. MiniLLM reduces this mismatch by optimizing reverse KL on student
samples \cite{gu2024minillm}, while generalized knowledge distillation
formalizes supervision on student-generated trajectories and includes OPD as
an on-policy instance \cite{agarwal2024onpolicy}. Recent studies show that
successful OPD also depends on student--teacher compatibility and token-level
utility, rather than teacher strength or discrepancy alone
\cite{li2026rethinking,xu2026tip}. REOPD builds on this on-policy setting but
targets the extrapolation residual rather than the complete distillation loss.

\paragraph{Reward Extrapolation for On-Policy Distillation.}
G-OPD casts OPD as dense KL-regularized reinforcement learning
\cite{yang2026learning}: the teacher--reference log-ratio serves as an
implicit reward, and the student--reference KL constrains policy deviation.
The formulation recovers OPD at $\lambda=1$ and defines ExOPD by
$\lambda>1$, where an additional teacher--reference residual is scaled by
$\lambda-1$. This objective-level extrapolation differs from weight-space
methods such as ExPO, which extrapolate model parameters after preference
optimization \cite{zheng2025model}. Although a suitable $\lambda$ can improve
distillation, fixed global scaling is sensitive to both log-ratio peaks and
the task domain. REOPD directly extends ExOPD by replacing its global
residual multiplier with online token- and micro-batch-level control.

\paragraph{Adaptive and Reliability-Aware Distillation.}
Adaptive distillation adjusts supervision according to training state or sample
quality. AdaKD adapts token selection and temperature
\cite{xie2026adakd}, while ASKD conditions a KL objective on externally
provided sample quality \cite{zhang2026askd}. Within OPD, TIP selects
informative alignment tokens \cite{xu2026tip}, and Prune-OPD attenuates
teacher supervision or truncates rollouts after local support drift
\cite{yang2026pruneopd}. These methods modify teacher alignment or the rollout
process; REOPD instead preserves alignment and controls only the
beyond-teacher residual.

Reliability can also be estimated from task-level feedback. SCOPE routes
verifier-labeled trajectories between self-reinforcement and OPD
\cite{zheng2026scope}; SG-OPD uses agreement between outcome and teacher
signals and verified teacher rollouts \cite{xu2026sgopd}; and reward-gated OPD
uses verifier feedback to regulate teacher guidance
\cite{akhondzadeh2026rewardgated}. REOPD addresses the complementary setting
without outcome labels or verifiers: it derives residual control solely from
student, teacher, and reference log-probabilities already computed in
white-box OPD.

\section{REOPD: Reliability-Adaptive Reward Extrapolation}
\label{sec:method}

\subsection{Problem Setup}

Let $\mathcal{D}$ denote the prompt distribution. Given a prompt $x\sim\mathcal{D}$, the student policy $\pi_\theta$ generates an on-policy response $y=(y_1,\ldots,y_T)$, where $y\sim\pi_\theta(\cdot\mid x)$. We denote the prefix state at token $t$ by $s_t=(x,y_{<t})$. The teacher and reference policies are denoted by $\pi_T$ and $\pi_{\mathrm{ref}}$, respectively. In the multi-teacher setting, $\pi_T$ denotes the domain teacher routed to the current example. Let $m_t\in\{0,1\}$ indicate whether $y_t$ is a valid response token.

For a sampled token $y_t$, we define the student--teacher alignment cost as

\begin{equation}
a_t =
\log\pi_\theta(y_t\mid s_t)
-\log\pi_T(y_t\mid s_t).
\label{eq:alignment-cost}
\end{equation}

Its expectation under the student distribution corresponds to the reverse KL divergence between the student and teacher at $s_t$. Thus, $a_t$ is an alignment cost term; the PPO advantage is defined later as the negative total token cost. G-OPD further defines the teacher--reference log-ratio as a dense implicit reward \cite{yang2026learning}:

\begin{equation}
r_t =
\log\pi_T(y_t\mid s_t)
-\log\pi_{\mathrm{ref}}(y_t\mid s_t).
\label{eq:implicit-reward}
\end{equation}

With a global reward coefficient $\lambda$, the sampled-token ExOPD cost can be written as

\begin{equation}
C_t^{\mathrm{ExOPD}}
=
a_t-(\lambda-1)r_t.
\label{eq:exopd-cost}
\end{equation}

When $\lambda=1$, Equation~\eqref{eq:exopd-cost} reduces to standard OPD. When $\lambda>1$, the residual term encourages the student to move beyond the teacher along the teacher--reference direction. However, the same multiplier $\lambda-1$ is applied to every token. This uniform scaling can amplify extreme implicit-reward peaks and requires $\lambda$ to be selected separately for different training settings. We therefore preserve $a_t$ and replace only the residual multiplier with a bounded, data-dependent coefficient.

\subsection{Method Overview}

Let $b$ denote a synchronized micro-batch whose sufficient statistics are aggregated across data-parallel ranks, and let $i$ index a response in $b$. REOPD constructs the effective extrapolation coefficient

\begin{equation}
\lambda_{b,i,t}
=
1+\gamma_b q_{b,i,t},
\label{eq:adaptive-lambda}
\end{equation}

where $q_{b,i,t}\in(0,1]$ is a token-level compatibility weight and $\gamma_b\in[0,\gamma_{\max}]$ is a shared micro-batch extrapolation budget. The resulting token cost is

\begin{equation}
C_{b,i,t}^{\mathrm{REOPD}}
=
a_{b,i,t}
-\gamma_b q_{b,i,t}r_{b,i,t}.
\label{eq:reopd-cost}
\end{equation}

Figure~\ref{fig:overview}(b) separates REOPD into a token path and a controller path. The token path retains the local residual $q_{b,i,t}r_{b,i,t}$, whereas the controller path all-reduces detached sufficient statistics to obtain the shared budget $\gamma_b$. Their product determines how much of the extra residual is applied to each token. Neither path modifies the teacher-alignment cost $a_{b,i,t}$.

This formulation contains several useful special cases. Setting $\gamma_b=0$ recovers OPD. Setting $q_{b,i,t}=1$ and $\gamma_b=\lambda-1$ recovers fixed-$\lambda$ ExOPD. A fixed $\gamma$ with adaptive $q_{b,i,t}$ gives token-only control, while $q_{b,i,t}=1$ with adaptive $\gamma_b$ gives micro-batch-only control.

\subsection{Token-Level Compatibility Weight}

REOPD estimates local student--teacher compatibility from log-probabilities already computed on sampled response tokens. We first construct the low-variance $k_3$ discrepancy proxy

\begin{equation}
\begin{aligned}
x_{b,i,t}
&=
\log\pi_T(y_{i,t}\mid s_{i,t})
-\log\pi_\theta(y_{i,t}\mid s_{i,t}),\\
\widehat{\delta}_{b,i,t}
&=
\exp(x_{b,i,t})-x_{b,i,t}-1.
\end{aligned}
\label{eq:compatibility-proxy}
\end{equation}

The compatibility weight is then defined as

\begin{equation}
q_{b,i,t}
=
\exp\left(
-\frac{\widehat{\delta}_{b,i,t}}{\tau}
\right),
\qquad \tau>0.
\label{eq:compatibility-weight}
\end{equation}

Because $\widehat{\delta}_{b,i,t}\geq0$, the weight lies in $(0,1]$ in exact arithmetic. A small sampled discrepancy gives $q_{b,i,t}\approx1$ and retains most of the extrapolation residual, whereas a large discrepancy yields a smaller weight. The temperature $\tau$ controls how rapidly this attenuation occurs.

We emphasize that $q_{b,i,t}$ measures local compatibility rather than task-level correctness. A small discrepancy does not guarantee that the teacher is correct, and a large discrepancy does not make its alignment signal useless. This is why REOPD applies $q_{b,i,t}$ only to the additional reward residual. The compatibility proxy and its resulting weight are detached from the computation graph. Intermediate log-ratios are numerically bounded in implementation, but no full-vocabulary KL or additional teacher forward pass is required.

\subsection{Micro-Batch Reliable Residual Statistics}

Token compatibility alone does not determine how much extrapolation the current micro-batch can support. REOPD therefore aggregates two statistics over valid response tokens. For compactness, let $\sum_b$ denote summation over all sequence--token pairs $(i,t)$ in synchronized micro-batch $b$. We define the compatibility-weighted residual proportion as

\begin{equation}
\rho_b
=
\frac{
\sum_b m_{i,t}|r_{b,i,t}|q_{b,i,t}
}{
\sum_b m_{i,t}|r_{b,i,t}|+\epsilon
}.
\label{eq:reliable-mass}
\end{equation}

The statistic $\rho_b\in[0,1]$ measures the fraction of residual magnitude retained after compatibility weighting. We further define the reliable residual scale

\begin{equation}
s_b
=
\left(
\frac{
\sum_b m_{i,t}
(q_{b,i,t}r_{b,i,t})^2
}{
\sum_b m_{i,t}+\epsilon
}
\right)^{1/2}.
\label{eq:reliable-scale}
\end{equation}

Thus, $\rho_b$ captures the relative amount of compatible residual, whereas $s_b$ captures its absolute RMS scale. All sufficient statistics are summed across data-parallel ranks before the ratios are evaluated, so each rank uses the same controller output.

To reduce micro-batch noise, REOPD maintains exponential moving averages:

\begin{equation}
\bar{z}_b
=
\beta\bar{z}_{b-1}
+(1-\beta)z_b,
\quad z\in\{\rho,s\}.
\label{eq:controller-ema}
\end{equation}

Both statistics and their moving averages are computed without gradient tracking.

\subsection{Bounded Micro-Batch Extrapolation Budget}

REOPD converts the smoothed statistics into a target extrapolation budget:

\begin{equation}
\widetilde{\gamma}_b
=
\mathrm{clip}
\left(
\frac{B_0\bar{\rho}_b}
{\bar{s}_b+\epsilon},
0,\gamma_{\max}
\right).
\label{eq:gamma-target}
\end{equation}

A larger $\bar{\rho}_b$ permits stronger extrapolation when a larger fraction of the residual remains compatible. In contrast, a larger $\bar{s}_b$ reduces the coefficient so that a micro-batch with large residual scale does not dominate the update. The upper bound $\gamma_{\max}$ provides an explicit limit on extrapolation.

The clipped target is further smoothed before being applied:

\begin{equation}
\gamma_b
=
\beta_\gamma\gamma_{b-1}
+(1-\beta_\gamma)\widetilde{\gamma}_b.
\label{eq:gamma-smooth}
\end{equation}

Since $\widetilde{\gamma}_b\in[0,\gamma_{\max}]$, the smoothed budget remains in the same interval when initialized within it. When Equation~\eqref{eq:gamma-target} is not clipped and $\epsilon$ is negligible, it approximately satisfies
$\widetilde{\gamma}_b\bar{s}_b
\approx B_0\bar{\rho}_b$.
This relation should be interpreted as an adaptive target rather than a strict constraint after clipping and smoothing.

The scale $B_0$ can be specified directly. In automatic mode, REOPD initializes it during the first $K_0$ controller updates from a scaled moving average of the alignment RMS,
$\kappa\,\mathrm{RMS}_b(a)$, and then keeps it fixed. An optional warm-up can hold $\gamma_b$ at a prescribed value during the initial training steps. The corresponding initialization length, smoothing coefficients, and warm-up configuration are reported with the training hyperparameters.

\subsection{Final Objective and Optimization}

The PPO-style actor update \cite{schulman2017ppo} uses the negative token
cost as its advantage:

\begin{equation}
A_{b,i,t}^{\mathrm{REOPD}}
=
-C_{b,i,t}^{\mathrm{REOPD}}.
\label{eq:reopd-advantage}
\end{equation}

In exact arithmetic, Equations~\eqref{eq:adaptive-lambda}--\eqref{eq:gamma-smooth} imply

\begin{equation}
1
\leq
\lambda_{b,i,t}
\leq
1+\gamma_{\max}.
\label{eq:lambda-bound}
\end{equation}

The advantage is masked to valid response tokens and passed to the same PPO-style policy surrogate used by the OPD baseline. The compatibility weights, sufficient statistics, moving averages, and extrapolation budget are all treated as stop-gradient control signals. REOPD therefore changes only the construction of the token advantage and reuses the student, teacher, and reference evaluations already required by G-OPD.

\begin{algorithm}[tb]
\caption{REOPD Training}
\label{alg:reopd}
\textbf{Input}: Prompt distribution $\mathcal{D}$; student
$\pi_\theta$; teacher $\pi_T$; reference $\pi_{\mathrm{ref}}$;
$\tau$, $\gamma_{\max}$, $\beta$, $\beta_\gamma$; optional $B_0$
or auto-calibration parameters $\kappa,K_0$\\
\textbf{Output}: Updated student policy $\pi_\theta$
\begin{algorithmic}[1]
\FOR{each synchronized micro-batch $b$}
    \STATE Sample responses $y\sim\pi_\theta(\cdot\mid x)$,
    $x\sim\mathcal{D}$
    \STATE Compute sampled-token log-probabilities under
    $\pi_\theta$, $\pi_T$, and $\pi_{\mathrm{ref}}$
    \STATE Compute $a$, $r$, $\widehat{\delta}$, and detached
    $q$ using Eqs.~\eqref{eq:alignment-cost}--\eqref{eq:compatibility-weight}
    \STATE All-reduce masked sufficient statistics across
    data-parallel ranks
    \STATE Compute $\rho_b$, $s_b$, and alignment RMS
    \STATE Update $B_0$ during the first $K_0$ controller calls
    if auto-calibration is enabled
    \STATE Update $\bar{\rho}_b$ and $\bar{s}_b$ using
    Eq.~\eqref{eq:controller-ema}
    \STATE Compute $\widetilde{\gamma}_b$ and $\gamma_b$ using
    Eqs.~\eqref{eq:gamma-target}--\eqref{eq:gamma-smooth}
    \STATE Construct $A^{\mathrm{REOPD}}$ using
    Eqs.~\eqref{eq:reopd-cost} and~\eqref{eq:reopd-advantage}
    \STATE Update $\theta$ with the PPO-style policy objective
\ENDFOR
\end{algorithmic}
\end{algorithm}

\section{Experiments and Analysis}
\label{sec:experiments}

We evaluate REOPD in single-teacher mathematics, single-teacher code
generation, and mixed-domain multi-teacher distillation. Our experiments are
designed to answer four questions: (i) whether REOPD improves over standard
on-policy distillation; (ii) whether it improves over fixed-coefficient ExOPD
at $\lambda=1.25$; (iii) how token-level
compatibility and micro-batch-level budgeting affect performance; and
(iv) how the adaptive controller evolves during training.

\subsection{Experimental Setup}
\label{sec:experimental-setup}

\paragraph{Models and data.}
We initialize both the student and the reference policy from
Qwen3-4B \cite{yang2025qwen3}. We use
task-specialized Qwen3-4B non-thinking policies trained with reinforcement
learning as the mathematics and code teachers. The mathematics run uses
57,046 level-6 examples from the filtered
DeepMath-103K training set \cite{he2025deepmath}, while the code run uses
25,276 examples from the Eurus code training split \cite{yuan2024eurus}.
The multi-teacher set contains 25,276 examples
from each domain. Each mixed-domain example carries a domain label that
routes its teacher log-probability to the corresponding mathematics or code
teacher.

\begin{figure*}[!t]
\centering
\includegraphics[width=\textwidth]{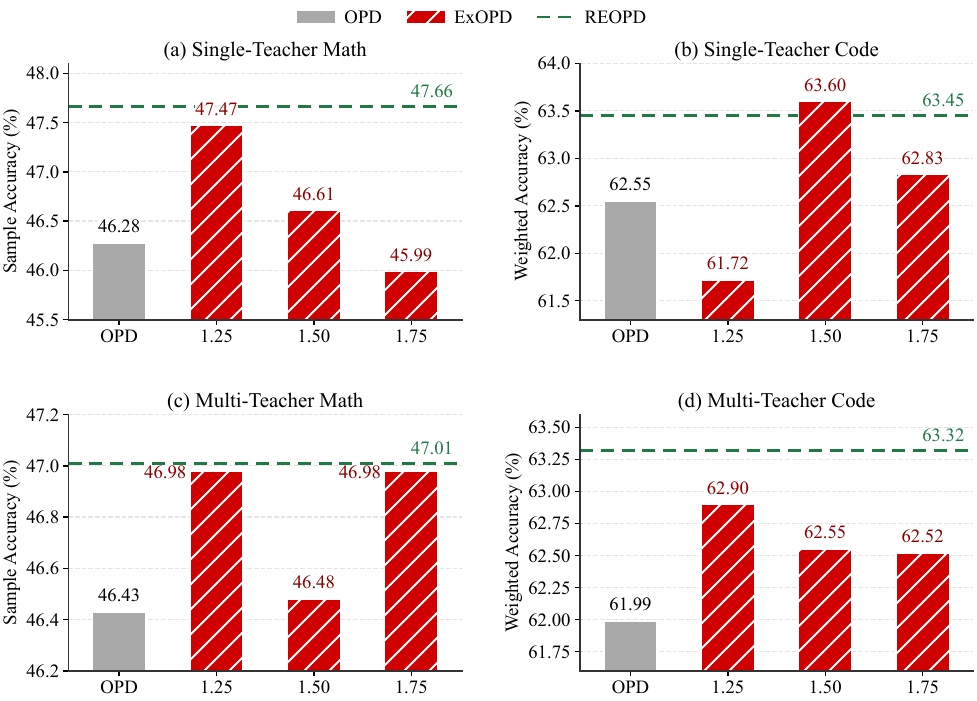}
\caption{Aggregate performance across fixed ExOPD coefficients. Gray bars
denote OPD, red hatched bars denote fixed ExOPD, and green dashed lines
denote REOPD. Axes are truncated to expose setting-dependent differences.}
\label{fig:lambda_sensitivity}
\end{figure*}

\paragraph{Baselines.}
We compare REOPD with standard OPD ($\lambda=1$) and fixed-coefficient ExOPD
at $\lambda=1.25$. Standard OPD performs teacher alignment without the
additional beyond-teacher residual, whereas ExOPD applies the same
extrapolation strength to every sampled token. We use the same
$\lambda=1.25$ baseline for mathematics and code in both the single- and
multi-teacher settings, without post-hoc coefficient selection. Sensitivity
results for other fixed coefficients are reported in
Fig.~\ref{fig:lambda_sensitivity}.

\begin{table*}[t]
\centering
\small
\setlength{\tabcolsep}{3.0pt}
\begin{tabular}{lccccccccc}
\toprule
& \multicolumn{5}{c}{Mathematical Reasoning} &
  \multicolumn{4}{c}{Code Generation} \\
\cmidrule(lr){2-6}\cmidrule(lr){7-10}
Method &
AIME24 & AIME25 & HMMT25-F & HMMT25-N & Avg. &
HE+ & MBPP+ & LCB & Avg. \\
\midrule
\multicolumn{10}{l}{\textit{Reference Models}} \\
Student &
23.65 & 22.50 & 12.50 & 9.27 & 16.98 &
80.34 & 64.88 & 17.43 & 56.83 \\
Teachers &
58.02 & 54.58 & 32.50 & 38.85 & 45.99 &
86.59 & 70.11 & 27.57 & 63.49 \\
\midrule
\multicolumn{10}{l}{\textit{Single-Teacher Distillation}} \\
OPD &
60.10 & 55.52 & 32.19 & 37.29 & 46.28 &
84.15 & 69.31 & 27.71 & 62.55 \\
ExOPD ($\lambda=1.25$) &
61.88 & \textbf{56.25} & 32.81 & \textbf{38.96} & 47.47 &
83.84 & 67.53 & \textbf{28.43} & 61.72 \\
REOPD &
\textbf{61.98} & 55.31 & \textbf{34.69} & 38.65 & \textbf{47.66} &
\textbf{85.67} & \textbf{70.57} & 27.29 & \textbf{63.45} \\
\midrule
\multicolumn{10}{l}{\textit{Multi-Teacher Distillation}} \\
OPD &
58.85 & \textbf{56.35} & 31.46 & \textbf{39.06} & 46.43 &
84.76 & 68.65 & 26.29 & 61.99 \\
ExOPD ($\lambda=1.25$) &
\textbf{62.92} & 56.04 & 31.98 & 36.98 & 46.98 &
\textbf{86.43} & 68.78 & \textbf{28.14} & 62.90 \\
REOPD &
61.35 & 55.10 & \textbf{33.75} & 37.81 & \textbf{47.01} &
85.82 & \textbf{70.11} & 27.57 & \textbf{63.32} \\
\bottomrule
\end{tabular}
\caption{Benchmark-level accuracy (\%). HMMT25-F/N denote the February/November
2025 contests, and LCB denotes the \texttt{test6} split of LiveCodeBench v6.
The mathematics average pools 3,840 completions; the code average is weighted
by task count over 2,868 completions. ExOPD uses $\lambda=1.25$; bold denotes
the best result within each distillation setting.}
\label{tab:main_results}
\end{table*}

\paragraph{Training details.}
All main comparisons use the actor checkpoint at global step 50. Each step
draws 1,024 prompts and one on-policy response per prompt. We optimize the
student with AdamW \cite{loshchilov2019decoupled} using a learning rate of
$10^{-5}$, weight decay $0.01$, gradient clipping at $1.0$, and one PPO
epoch. The maximum prompt and response
lengths are 2,048 and 16,384 tokens, respectively. Training uses BF16 mixed
precision with FSDP on eight NVIDIA A100 80GB GPUs, and rollout generation
uses tensor parallelism of size four.

For REOPD, we set $\tau=0.007$, $\gamma_{\max}=1$, $\kappa=0.5$,
$\beta=0.95$, and $\beta_\gamma=0.9$. The initial budget scale $B_0$ is
automatically calibrated over the first ten controller calls. The mathematics
run uses a five-call warm-up with $\gamma=0.25$; code and multi-teacher runs
use no warm-up. The multi-teacher setting maintains one shared controller
across the two domains. Unless otherwise stated, all methods use the same
student initialization, teacher, training file, prompt budget, and step-50
checkpoint. Equal prompt budgets do not imply equal generated-token budgets
because response lengths differ across methods.

\paragraph{Ablation protocol.}
Component ablations are conducted in the single-teacher mathematics setting.
All variants use the same mathematics training set, step-50
checkpoint, and evaluation protocol with seed 42. The \texttt{no\_q} variant
sets $q_t=1$ while retaining the adaptive micro-batch budget and its bound.
The \texttt{no\_bound} variant retains $q_t$ and the adaptive controller but
removes only the explicit upper bound on $\gamma_b$, while preserving the
controller statistics, synchronization, EMA, and update frequency.
The \texttt{no\_batch} variant retains $q_t$ but replaces $\gamma_b$ with a
fixed $\gamma=\lambda_0-1$. To measure its dependence on this fixed choice,
we evaluate $\lambda_0\in\{1.0,1.25,1.5,1.75\}$.

\paragraph{Evaluation.}
For mathematics, we evaluate on AIME 2024, AIME 2025, HMMT February 2025, and
HMMT November 2025. Each benchmark contains 30 problems, and we sample 32
responses per problem with temperature 1 and top-$p=1$. We report pooled
sample accuracy over all 3,840 completions.
For code, we evaluate HumanEval+ (164 tasks) and MBPP+ (378 tasks) using
EvalPlus \cite{chen2021codex,austin2021program,liu2023evalplus}, together
with the 175-task \texttt{test6} split of LiveCodeBench
v6 \cite{jain2024livecodebench}. We draw four responses per
problem using the same sampling temperature and report official pass@1. Our
aggregate code score is the task-count-weighted accuracy over all 2,868
completions, rather than the unweighted mean of the three benchmark
percentages. No external judge is used.

\subsection{Main Results}
\label{sec:main-results}

\paragraph{Single-teacher distillation.}
As shown in Table~\ref{tab:main_results}, REOPD reaches 47.66\% pooled sample
accuracy on mathematics, exceeding both OPD at 46.28\% and ExOPD at
$\lambda=1.25$ at 47.47\%. On code generation, REOPD obtains 63.45\%
weighted accuracy and remains comparable to G-OPD; under the common
$\lambda=1.25$ baseline, it exceeds ExOPD at 61.72\% and OPD at 62.55\%.

\paragraph{Multi-teacher distillation.}
When mathematics and code examples share one student and one REOPD
controller, REOPD reaches 47.01\% mathematics sample accuracy and 63.32\%
code weighted accuracy. Both results exceed OPD at 46.43\% and 61.99\%, as
well as ExOPD at $\lambda=1.25$ at 46.98\% and 62.90\%, respectively. These
results show that a shared adaptive controller can improve both domains
without selecting separate fixed coefficients for mathematics and code.

\subsection{Sensitivity to a Global Extrapolation Coefficient}
\label{sec:lambda-sensitivity}

Figure~\ref{fig:lambda_sensitivity} exposes substantial variation in the
preferred fixed coefficient. Single-teacher mathematics peaks at
$\lambda=1.25$ with 47.47\%, whereas single-teacher code peaks at
$\lambda=1.5$ with 63.60\%. In multi-teacher distillation,
$\lambda=1.25$ and $1.75$ tie on mathematics at 46.98\%, while
$\lambda=1.25$ is best on code at 62.90\%. Performance is non-monotonic:
increasing $\lambda$ beyond its setting-specific optimum can reduce
accuracy.

To test whether REOPD's adaptation can recover the effect of tuning a fixed
coefficient for each setting, we compare it directly with the best ExOPD
result from each G-OPD coefficient sweep. REOPD changes accuracy by $+0.19$,
$-0.15$, $+0.03$, and $+0.42$ percentage points on single-teacher
mathematics, single-teacher code, multi-teacher mathematics, and
multi-teacher code, respectively. It therefore exceeds the corresponding
best fixed-coefficient result in three of the four settings and trails it by
only 0.15 points on single-teacher code. Overall, REOPD reaches comparable or
better performance than the task-specific best fixed coefficient, showing
that token-level compatibility and micro-batch budgeting can adapt the
extrapolation strength effectively without a separate $\lambda$ sweep for
each setting.

Taken together, the best fixed coefficient varies across task domains and
teacher configurations, whereas REOPD replaces per-setting selection of a
global $\lambda$ with online adaptation. REOPD is not hyperparameter free:
the controller retains hyperparameters shared across settings, and the
mathematics run uses the warm-up schedule described in
Section~\ref{sec:experimental-setup}.

\subsection{Ablation Studies}
\label{sec:ablations}

\begin{table}[t]
\centering
\small
\setlength{\tabcolsep}{3.2pt}
\begin{tabular}{lcccr}
\toprule
Variant & Token $q$ & Adapt.\ $\gamma_b$ & $\lambda_0$ & Acc. \\
\midrule
Full REOPD & Yes & Yes & -- & 47.66 \\
\texttt{no\_q} & -- & Yes & -- & 43.39 \\
\texttt{no\_bound} & Yes & Yes & -- & 47.16 \\
\midrule
\texttt{no\_batch} & Yes & -- & 1.00 & 46.48 \\
\texttt{no\_batch} & Yes & -- & 1.25 & \textbf{47.66} \\
\texttt{no\_batch} & Yes & -- & 1.50 & 46.81 \\
\texttt{no\_batch} & Yes & -- & 1.75 & 46.68 \\
\bottomrule
\end{tabular}
\caption{Component ablations on single-teacher mathematics only.
\texttt{no\_q} sets $q_t=1$; \texttt{no\_bound} removes the upper bound on
$\gamma_b$; and \texttt{no\_batch} fixes $\gamma=\lambda_0-1$. Accuracy pools
3,840 completions. Bold marks the best \texttt{no\_batch} result.}
\label{tab:ablation}
\end{table}

\begin{figure*}[!t]
\centering
\includegraphics[width=0.90\textwidth]{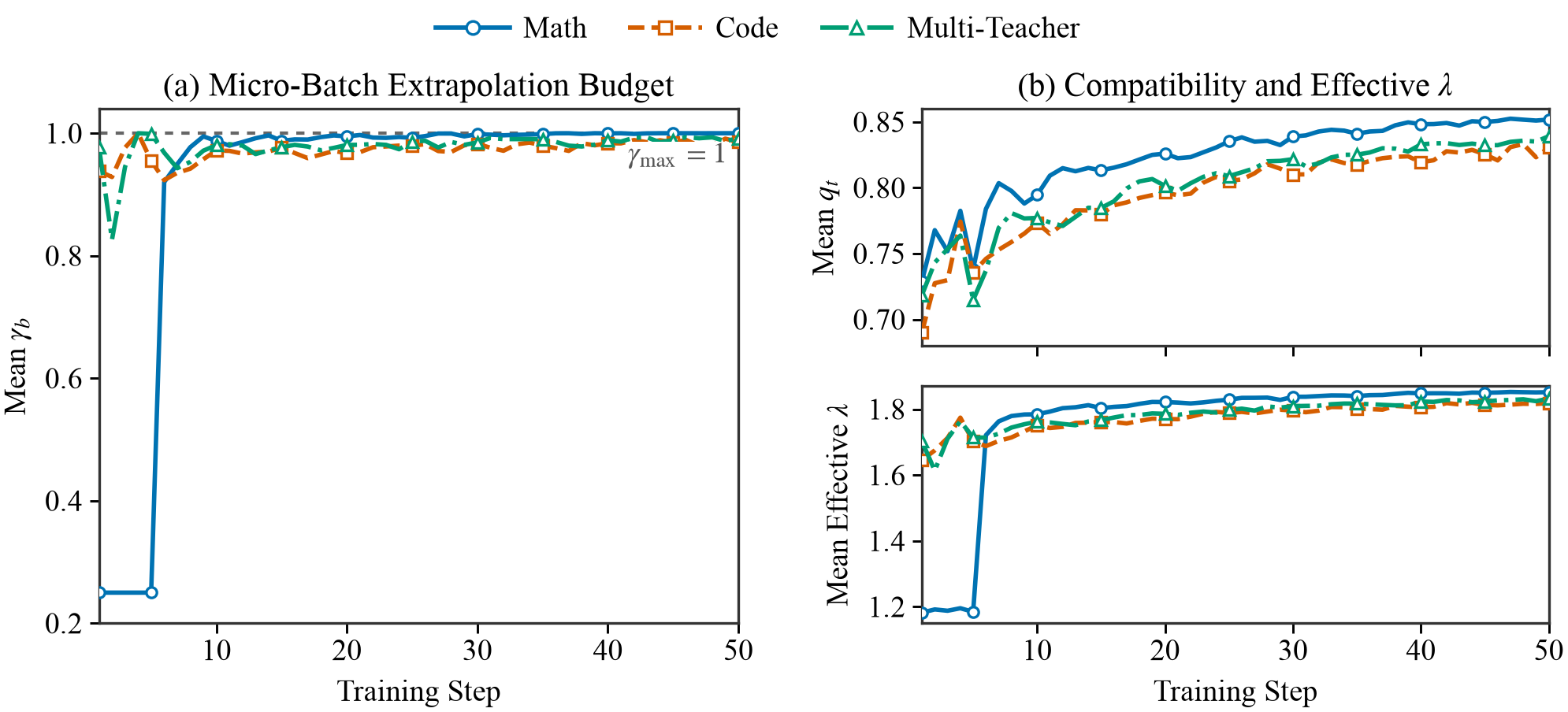}
\caption{Raw controller trajectories. Panel (a) reports the micro-batch
budget $\gamma_b$; panel (b) reports mean compatibility $q_t$ and effective
coefficient $\lambda_{b,i,t}=1+\gamma_bq_{b,i,t}$.}
\label{fig:controller_dynamics}
\end{figure*}

Table~\ref{tab:ablation} isolates the two adaptation levels in the
single-teacher mathematics setting. Removing token compatibility
(\texttt{no\_q}) reduces pooled accuracy from 47.66\% to 43.39\%, a drop of
4.27 percentage points. The decrease occurs on all four benchmarks
(2.40--7.19 points). Thus, applying a shared micro-batch budget uniformly to
all tokens is insufficient; token compatibility is the central component of
REOPD.

The \texttt{no\_bound} variant isolates the explicit budget cap while
preserving both levels of adaptation. It obtains 47.16\% pooled accuracy
(1,811/3,840), 0.50 percentage points below Full REOPD. Its benchmark
accuracies are 61.04\% on AIME 2024, 53.96\% on AIME 2025, 34.58\% on HMMT
February 2025, and 39.06\% on HMMT November 2025. Relative to Full REOPD,
these scores are lower by 0.94, 1.35, and 0.11 points on the first three
benchmarks, respectively, but higher by 0.41 points on HMMT November 2025.
The cross-benchmark effect is therefore mixed, but the lower pooled accuracy
suggests that the explicit bound provides a modest safeguard; its contribution
is substantially smaller than that of token compatibility.

The \texttt{no\_batch} sweep tests the complementary question by retaining
$q_t$ while using a fixed base coefficient $\lambda_0$. Its pooled accuracy
ranges from 46.48\% to 47.66\%. The best configuration uses
$\lambda_0=1.25$ and reaches 47.66\%, matching Full REOPD. Thus, adaptive
micro-batch budgeting matches the best tuned token-only variant while
avoiding a setting-specific $\lambda_0$ sweep. Together with the
\texttt{no\_q} and \texttt{no\_bound} results, the ablation identifies
token-level compatibility as the key component and the explicit bound as a
modest safeguard, while the online micro-batch budget provides adaptive
control without fixed-coefficient selection.

\subsection{Controller Dynamics}
\label{sec:controller-analysis}

The logged trajectories confirm that REOPD does not use a constant effective
coefficient. Over the first ten steps, the mean budget $\gamma$ is 0.608,
0.953, and 0.957 for mathematics, code, and multi-teacher training,
respectively; over the final ten steps, these values increase to 1.000,
0.986, and 0.990. Meanwhile, mean token compatibility increases from
0.774/0.745/0.754 to 0.850/0.827/0.834, and the mean effective coefficient
increases from 1.477/1.710/1.720 to 1.850/1.815/1.826. The controller
therefore adapts most strongly early in training. Because the micro-batch
budget approaches its upper bound later, the remaining late-stage variation
is primarily supplied by the token-level compatibility weight.

\section{Conclusion and Limitations}
\label{sec:conclusion}

We studied reward extrapolation for on-policy distillation, where a fixed
global coefficient applies the same beyond-teacher residual gain to every
sampled token and must be selected separately for different training
settings. REOPD replaces this global multiplier with a token-level
compatibility weight and a bounded micro-batch extrapolation budget. It
preserves standard teacher alignment, adapts only the additional
teacher--reference residual, and requires no verifier or extra rollout.
Across repeated experiments, REOPD outperforms G-OPD on single-teacher
mathematics and on both multi-teacher domains, while achieving comparable
performance on single-teacher code. In the mathematics ablation, removing
$q_t$ lowers accuracy by 4.27 points, removing the explicit budget bound
lowers accuracy by 0.50 points, and the best token-only variant at
$\lambda_0=1.25$ matches Full REOPD. These results identify token-wise
residual filtering as the principal component, with the explicit bound
providing a modest safeguard and online micro-batch budgeting preserving
tuned performance without a setting-specific coefficient sweep.

Several limitations remain. First, the compatibility weight is a sampled
student--teacher discrepancy proxy rather than a correctness estimator; it
cannot identify trajectories on which the student and teacher agree but are
both wrong. Second, REOPD retains controller choices such as $\tau$,
$\gamma_{\max}$, and the
calibration of $B_0$; its statistics depend on micro-batch composition, and
the budget approaches its upper bound late in training. Future work should
evaluate additional model families, scales, and teacher configurations, and
compare shared versus per-teacher controllers to further test generality.

\bibliography{references}


\end{document}